\documentclass{article}
\usepackage{ijcai24}

\usepackage{times}
\usepackage{soul}
\usepackage{url}
\usepackage[hidelinks]{hyperref}
\usepackage[utf8]{inputenc}
\usepackage[small]{caption}
\usepackage{graphicx}
\usepackage{amsmath}
\usepackage{amsthm}
\usepackage {amssymb}
\usepackage{bm}
\usepackage{booktabs}
\usepackage{algorithm}
\usepackage{algorithmic}
\usepackage{color}
\usepackage{enumitem}
\usepackage{multirow}
\usepackage[switch]{lineno}

\title{Towards Generalizable Neural Solvers for Vehicle Routing Problems\\ via Ensemble with Transferrable Local Policy}

\author{
    Author Name
    \affiliations
    Affiliation
    \emails
    email@example.com
}

\author{
Chengrui Gao$^{1,2}$
\and
Haopu Shang$^{1,2}$\and
Ke Xue$^{1,2}$\and
Dong Li$^3$\And
Chao Qian$^{1,2}$\\
\affiliations
$^1$National Key Laboratory for Novel Software Technology, Nanjing University, China\\
$^2$School of Artificial Intelligence, Nanjing University, China\\
$^3$Huawei Noah’s Ark Lab
\emails
\{gaocr, shanghp, xuek\}@lamda.nju.edu.cn,
lidong106@huawei.com,
qianc@nju.edu.cn
}

\begin{document}

\maketitle

\begin{abstract}
    Machine learning has been adapted to help solve NP-hard combinatorial optimization problems. One prevalent way is learning to construct solutions by deep neural networks, which has been receiving more and more attention due to the high efficiency and less requirement for expert knowledge. However, many neural construction methods for Vehicle Routing Problems~(VRPs) focus on synthetic problem instances with specified node distributions and limited scales, leading to poor performance on real-world problems which usually involve complex and unknown node distributions together with large scales. 
    To make neural VRP solvers more practical, we design an auxiliary policy that learns from the local transferable topological features, named local policy, and integrate it with a typical construction policy (which learns from the global information of VRP instances) to form an ensemble policy. With joint training, the aggregated policies perform cooperatively and complementarily to boost generalization. The experimental results on two well-known benchmarks, TSPLIB and CVRPLIB, of travelling salesman problem and capacitated VRP show that the ensemble policy significantly improves both cross-distribution and cross-scale generalization performance, and even performs well on real-world problems with several thousand nodes. 
\end{abstract}

\section{Introduction}

Vehicle Routing Problems (VRPs) are a kind of canonical NP-hard combinatorial optimization problems, with broad applications in logistics~\cite{logistics}. They have been widely studied for decades, and various heuristic methods are proposed, such as LKH~\cite{LKH2,LKH3}, SISR~\cite{SISR}, and HGS~\cite{HGS}. However, existing heuristic methods are highly domain-specific, require lots of expert knowledge, and suffer from low efficiency in solving large-scale instances, which limit their application.

Recently, machine learning has been adapted to solve combinatorial optimization problems~\cite{bengio_review,cappart2023combinatorial}, which utilizes deep neural networks to extract features, learn heuristics, and generate solutions, named Neural Combinatorial Optimization~(NCO)~\cite{NCO_RL}. 
Depending on the training scheme, NCO includes Supervised Learning~(SL)~\cite{gnn_2,NIE} and Reinforcement Learning~(RL) methods~\cite{AM}. 
SL-based methods usually require numerous (nearly) optimal solutions as labels, which are hardly applicable if optimal solutions are expensive to obtain~\cite{NCO_RL}. 
RL-based methods train the model using objective function evaluations as rewards, which are effective in most scenarios~\cite{JSSP,ELS}. 

According to the process of generating solutions, RL-based methods can be generally divided into improvement and construction methods. 
Improvement methods learn to iteratively improve the quality of solutions, e.g., by selecting appropriate operators~\cite{L2I}.
They have the advantage of exploiting human knowledge, but suffer from limited search efficiency~\cite{neural-kopt} and high latency in inference~\cite{difusco}. 
Construction methods learn to extend a partial solution sequentially until a complete solution is constructed. 
This procedure can be modeled as a Markov Decision Process (MDP) and solved by RL~\cite{NCO_RL,s2v_dqn}. 
Construction methods are very computationally efficient, which can solve hundreds of instances in a few seconds. 
However, most construction methods only work well on problems with simple node distributions and limited scales. Their performance deteriorates dramatically on cross-distribution and cross-scale problems that are common in practice~\cite{rethink,DRO}. 

Towards generalizing to cross-distribution and cross-scale problem instances, several methods have been proposed recently. 
Some of them focus on generalization to large-scale problems. They often utilize the divide-and-conquer framework that can extend small-scale models to generalizable ones~\cite{gnn_MCTS_2,real-time,SO}. 
Meanwhile, the inductive bias of symmetries~\cite{BQ-NCO} and some novel neural architectures, such as heavy decoder~\cite{LEHD} and diffusion models~\cite{difusco}, are found to be useful for cross-scale generalization. The other methods aim to address the cross-distribution issue. A simple solution is directly training the model on instances with various node distributions, which is effective but can be further improved by applying some advanced techniques, 
such as distributionally robust optimization~\cite{DRO}, knowledge distillation~\cite{AMDKD}, graph contrastive learning~\cite{gcl}, and meta-learning~\cite{meta-RL,meta,meta-sage}. Note that the latest meta-learning method~\cite{omni-vrp} considers an omni setting including both cross-distribution and cross-scale generalization. 

In this paper, we also focus on the challenging omni-generalization setting to pursue better performance on real-world problems which often have complex and unknown node distributions together with large scales. 
Unlike previous methods that rely on the divide-and-conquer strategy or try to improve learning algorithms, 
we propose an ensemble method that integrates a policy of learning from global information of VRPs and a local policy with transferability, where their strengths are combined to boost generalization. 

Considering that the general optimization objective of VRPs is to achieve the shortest routing length, the local neighborhood, containing adjacent nodes, plays an important role in the decision of moving from node to node. Meanwhile, local topological features have great potential to be transferable across both node distributions~\cite{gcl} and scales. To leverage the properties of local information, we propose a novel local policy that restricts the state and action space to a small number of local neighbor nodes. Furthermore, we integrate the local policy with a typical construction policy (e.g., POMO~\cite{POMO}) called global policy, which learns from the global information of complete VRP instances, to form an ensemble policy. The local and global policies are jointly trained to perform cooperatively and complementarily, achieving a good generalization. 

We conduct experiments on two typical VRPs: Travelling Salesman Problem~(TSP) and Capacitated VRP~(CVRP). The proposed ensemble policy is trained on synthetic small-scale problem instances with uniform node distribution. 
Instead of using test datasets with fixed node distribution and limited problem scale, we select TSPLIB~\cite{TSPLIB} and CVRPLIB~\cite{vrplib_x} benchmarks to evaluate the generalization performance, which contain diverse instances with complex node distributions and large scales. 
The empirical results show that the ensemble policy significantly improves both cross-distribution and cross-scale generalization performance, and can perform better than state-of-the-art construction methods (e.g., BQ~\cite{BQ-NCO} and LEHD~\cite{LEHD}) in most cases. Moreover, the proposed method even performs well on real-world problems~\cite{real-world} with several thousand nodes, while most construction methods can hardly solve such real-world problems directly. 
The ablation study also verifies the key role of the transferrable local policy for better generalization.
\section{Background}
\begin{figure}[ht]
    \begin{center}
        \includegraphics[width=0.98\linewidth]{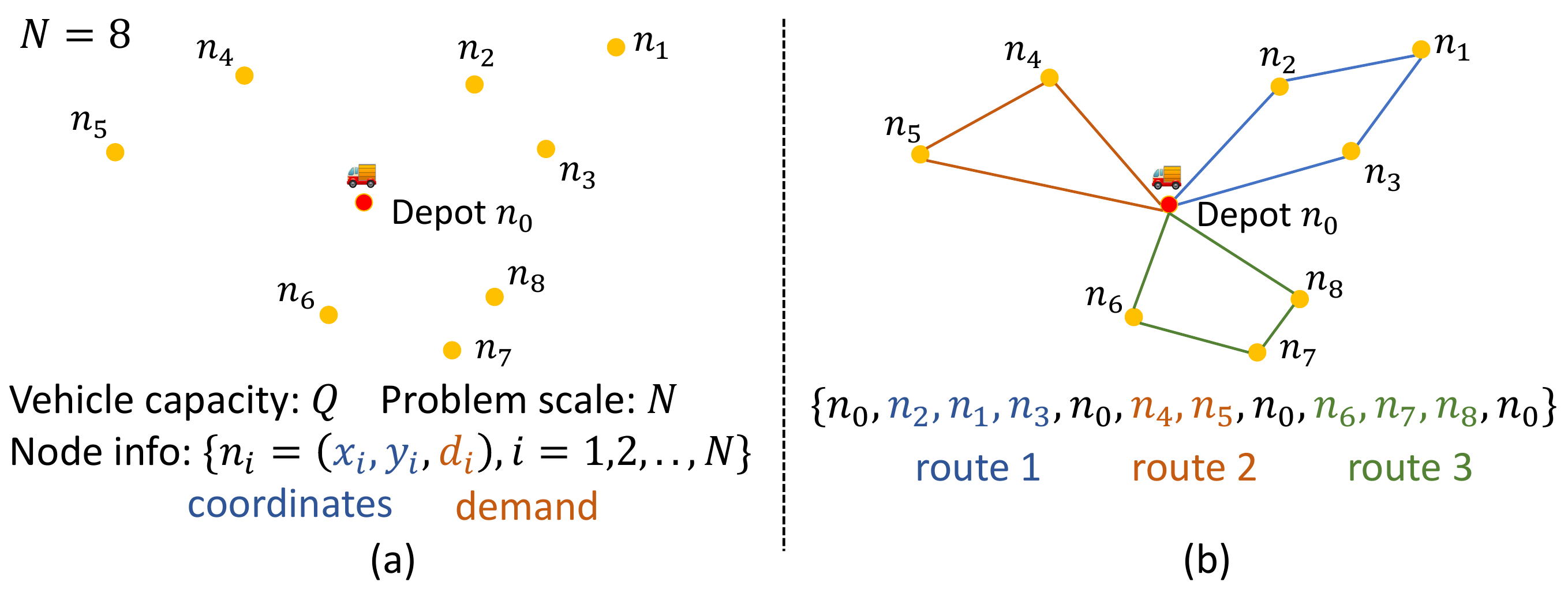}
        \caption{Example illustration of CVRP. (a) A CVRP instance and the information of its nodes and vehicles. (b) The sequence representation of a CVRP solution. }
        \label{vrp}    
    \end{center}\vspace{-1.0em}
\end{figure}
\subsection{Vehicle Routing Problems}
VRPs~\cite{dantzig1959truck} are well-known NP-hard combinatorial optimization problems with wide applications~\cite{logistics}. In this paper, we consider two typical VRPs, i.e., Euclidean TSP and CVRP. Given a CVRP instance $I$ as shown in Figure~\ref{vrp}(a), it contains a depot node $n_0$ and $N$ customer nodes $\{n_1,n_2,\ldots,n_N\}$ to serve by multiple identical vehicles, where each node $n_i$ has static Euclidean coordinate $(x_{n_i}, y_{n_i})$ and demand $d_i$, and the vehicles have identical capacity $Q$. The vehicles start from the depot node, visit all the customer nodes, and fulfill their demands. Every customer node must be visited exactly once. If a vehicle runs out of capacity, it has to return to the depot node. The objective of CVRP is to minimize the total travelling distance (Euclidean distance) of the vehicles, i.e., 
\begin{equation*}
    f(\bm{\tau}) = \sum_i \sqrt{(x_{\tau_i} - x_{\tau_{i+1}})^2 + (y_{\tau_i} - y_{\tau_{i+1}})^2},
\end{equation*}
without violating the constraints of no repetitive visiting and limited capacity.
As shown in Figure~\ref{vrp}(b), an example solution of CVRP can be naturally represented by a sequence $\bm{\tau}=\{\tau_0,\tau_1,\ldots\}$ (where $\tau_i\in \{n_0,n_1,\ldots,n_N\}$), which indicates the visiting order of nodes. The Euclidean TSP can be regarded as a simplified version of CVRP that only has one route and no capacity constraint. 

\subsection{Neural Construction Methods}
\label{section:construction}
Neural construction methods learn to construct the sequential solution in an autoregressive way~\cite{PN}, i.e., learn the probability distribution $P(\bm{\tau})$ of generating a solution $\bm{\tau}$ as
\begin{equation*}
    P(\bm{\tau}) = \Pi_t \, P(\tau_t \mid I,\{\tau_0, \tau_1, ... , \tau_{t-1}\}).
\end{equation*}
This construction procedure can be naturally formulated as a Markov Decision Process~(MDP) that takes the partial solution $\{\tau_0, \tau_1, ..., \tau_{t-1}\}$ as state and the next node $\tau_t$ to visit as action. 
Following such MDP, RL can be used to solve VRPs~\cite{NCO_RL,AM}. In practice, the partial solution information contained in the state is always reduced to the current and start nodes. 

Recent prevalent construction methods often employ Transformer~\cite{attention} without positional embedding to encode unordered nodes and decode solutions, including Attention Model~(AM)~\cite{AM} and Policy Optimization with Multiple Optima~(POMO)~\cite{POMO}. Their models have an encoder-decoder architecture. In brief, the encoder learns the node embeddings from the complete VRP graph that covers the global information of a problem instance, and the decoder computes the compatibility of query and keys, i.e.,  
\begin{equation*}
    \bm{u}_{\mathrm{global}} = \frac{\bm{q}^{\mathrm{T}} \mathbf{K}}{\sqrt{d}}, \,\bm{q}\in \mathbb{R}^{d\times 1}, \mathbf{K}\in \mathbb{R}^{d \times N},
\end{equation*}
as the score for selecting actions,
where $\bm{q}$ represents the query vector, $\mathbf{K}$ represents the matrix of keys, $N$ is the number of customer nodes, and $d$ is the embedding dimension. 

The query vector $\bm{q}$ comes from the context embedding that contains the information of the current node and the loaded capacity, representing the state. 
The keys $\mathbf{K}$ are the embeddings of candidate nodes that represent actions. Thus, the score $\bm{u}_{\mathrm{global}}$ measures the compatibility of state and actions. 
To avoid violating constraints, $\bm{u}_{\mathrm{global}}$ is masked as $\bm{u}_{\mathrm{masked}}$ with the $i$-th dimension
\begin{equation*}
    u_{\mathrm{masked}}^i = \left\{
        \begin{aligned}
     C\cdot \tanh(u_{\mathrm{global}}^i), & \quad \text{if node $n_i$ is valid}, \\
     -\infty, & \quad \text{otherwise}, \\
    \end{aligned}
    \right.
\end{equation*} 
where $\tanh$ clips scores to be within $[-1, 1]$, $C$ is a parameter to control the scale, and the scores of invalid actions are set as $-\infty$. The final action probability is computed as
\begin{equation*}
    \bm{\pi}_{\mathrm{global}} = \text{softmax}(\bm{u}_{\mathrm{masked}}). 
\end{equation*}
In each step, the decoder selects one node to grow the partial solution and transits to a new state. 
The decoder repeats until a complete solution is constructed. 
When the construction is done, the negative of the objective function is computed as the reward $R(\bm{\tau})=-f(\bm{\tau})$. A prevalent RL algorithm is REINFORCE~\cite{reinforce} with greedy rollout baseline~\cite{AM} or shared baseline~\cite{POMO}. 

\subsection{Generalization Issue}

Most existing methods are trained on small-scale problem instances (where $N\le 100$) generated from uniform node distribution, and tested on instances from the \emph{same} distribution. However, real-world problems are often with complex and unknown node distributions and large scales. 
This has led to the less satisfactory generalization performance of existing neural methods, which is recognized as a bottleneck for applications~\cite{rethink,AMDKD,how_good}. Many related works have tried to alleviate the generalization issue by leveraging divide-and-conquer strategy~\cite{real-time}, meta learning~\cite{meta}, symmetry of CO problems~\cite{BQ-NCO}, and heavy decoder architecture~\cite{LEHD}, while most of them ignore either the cross-distribution or scale setting. 

\begin{figure*}
    \begin{center}
        \includegraphics[width=0.98\linewidth]{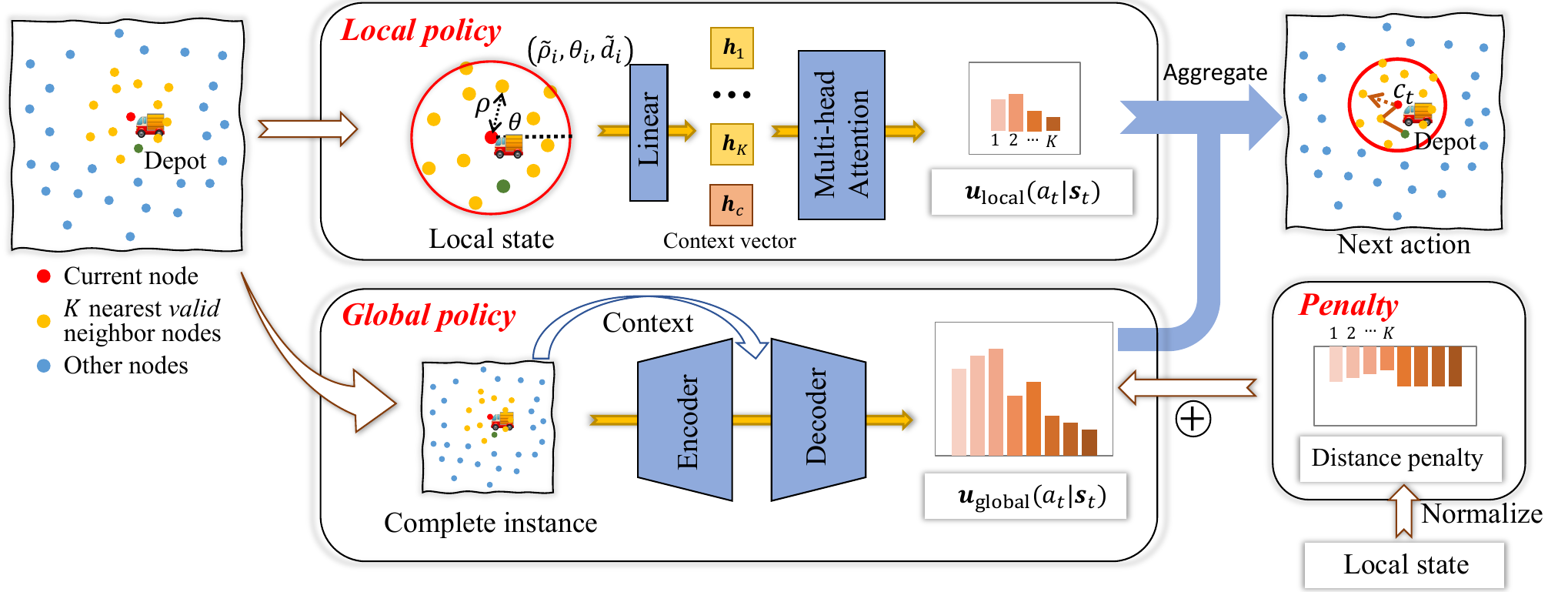}
        \caption{Framework of the ELG method, which aggregates a transferrable local policy that focuses on a small neighborhood of the current node and a global policy that learns from the information of the complete instance. }
        \label{framework}
    \end{center}\vspace{-1.0em}
\end{figure*}

In this paper, we aim to address the generalization issue by ensemble with a transferrable local policy. Unlike a related ensemble method for NCO~\cite{ensemble} that employs the well-known Bootstrap method, we leverage different
state and action spaces (i.e., local and global views) to generate diverse and complementary policies. 
To effectively verify the generalization performance, we select the challenging CVRPLIB~\cite{vrplib_x,real-world} and TSPLIB~\cite{TSPLIB} as our test datasets since they cover diverse instances and many of them are from the real world, while previous synthetic test datasets with either fixed node distributions or limited problem scales can only cover a narrow subspace of the whole problem space. 

\section{Method}
This section presents the proposed \textbf{E}nsemble of \textbf{L}ocal and \textbf{G}lobal policies~(ELG), including transferrable local policy, neural architecture, and ensemble strategy. Overall, the proposed ELG contains two base policies, i.e., a primary global policy and an auxiliary local policy, as shown in Figure~\ref{framework}. The primary global policy can be any popular neural combinatorial optimization model that learns from the global information of a complete VRP instance. 
Specifically, we choose POMO as the global policy, which has been introduced in Section~\ref{section:construction}. The process of constructing solutions of the local policy is similar to that of the global policy, but the state and action space is different, which focuses on the local neighborhood of the current node. 
\subsection{Transferrable Local Policy}
When solving VRPs, a common observation is that the optimal action (i.e., next node) is usually included in a small local neighborhood of the current node~\cite{LKH2}, and the patterns of such local neighborhoods have the potential to be transferrable across various node distributions~\cite{gcl} and scales. Inspired by these facts, we propose to design a local MDP formulation that restricts the state and action space to the local neighborhood, aiming to achieve better generalization performance. 
The overall procedure of the proposed local policy is illustrated in Figure~\ref{framework}. Here we define the state and action space of local policy in details. 
\paragraph{State.} The state space is reduced to a small local neighborhood $\mathcal{N}_K(c_t)$, which contains the $K$ nearest \textit{valid} neighbor nodes of the current node $c_t$. 
The \textit{valid} nodes are unvisited ones that meet constraints, e.g., capacity constraint in CVRP. 
Intuitively, the local state represents a subproblem that the policy needs to solve in the next few steps. By learning to solve local subproblems, the policy captures more intrinsic characteristics of VRPs, which are transferable across diverse problem instances. 
To better represent the features of the local state, we utilize polar coordinates $(\rho_i, \theta_i)$ centered at the current node $c_t$ to indicate the location information of the neighbor nodes, where $i$ is the index of node $n_i\in \mathcal{N}_K(c_t)$. 
The polar coordinates directly provide the relative distances~(i.e., edge cost) to $c_t$, and we can normalize all the $\rho_i$ into $[0, 1]$ by $ \tilde{\rho}_i = \rho_i/\max\{\rho_i\mid n_i\in \mathcal{N}_K(c_t)\}$. 
Therefore, all the neighbor nodes are in a unit ball. 
Such local topological features are insensitive to the changes in node distribution and problem scale.
The state of CVRP also contains the node demands $\{\tilde{d}_i\mid n_i\in \mathcal{N}_K(c_t)\}$, normalized by the remaining capacity $Q_{\text{remain}}$, i.e., $\tilde{d}_i = d_i/Q_{\text{remain}}$. Thus, $Q_{\text{remain}}$ is no longer required to be included in the state. 

\paragraph{Action.} The local policy outputs the scores $\bm{u}_{\mathrm{local}}$ for selecting the next node to visit from the local neighbors, where the score of node $n_i$ can be formulated as
\begin{equation*}
    u^i_{\mathrm{local}} = \left\{
    \begin{aligned}
        (g_{\bm{\theta}}(\bm{s}_t))_i, & \;\;\text{if} \; n_i \in \mathcal{N}_K(c_t), \\
    0, & \;\;\text{otherwise},\\ 
    \end{aligned}
    \right.
\end{equation*}
where $\bm{s}_t=\{[\tilde{\rho}_i, \theta_i, \tilde{d}_i]\mid  n_i\in \mathcal{N}_K(c_t)\}$ for CVRP~(note that demands $\tilde{d}_i$ are removed for TSP),
and $g_{\bm{\theta}}$ is a parameterized neural network, which will be described in Section~\ref{section:local_net}. 

The transition and reward design remains the same as the global policy~\cite{POMO}. 
Overall, the local policy learns from the local topological features and outputs the probability of selecting neighbor nodes. 
The advantages of the proposed local policy can be summarized as follows: 1) The local pattern is transferrable since it is insensitive to the shifts of node distribution and problem scale; 2) The local MDP gives a good inductive bias that the policy can focus on the local neighbor nodes which are probably more promising. 

\subsection{Neural Architecture}
\label{section:local_net}
To parameterize the local policy, we design an attention-based neural network, which has been shown to be effective for solving VRPs~\cite{AM,POMO}. 
To ensure the efficiency of local policy, we simply use a lightweight architecture, which consists of one shared linear layer for node embedding and one multi-head attention layer~\cite{attention} for decoding action scores. 

The node features of local neighbors $\{\bm{x}_i=[\tilde{\rho}_i, \theta_i, \tilde{d}_i]\mid n_i\in \mathcal{N}_K(c_t)\}$ are firstly mapped into $d_e$-dimension embeddings using a shared linear layer with parameters $[W;\bm{b}]$. 
Unlike previous attention models~\cite{AM,POMO} that abandon positional encoding to acquire permutation-invariant node embeddings, we retain the positional encoding to incorporate distance ordering information into the embeddings, which can benefit the local node selection. 
Concretely, the neighbor nodes $\bm{x}_i$ are sorted in ascending order by relative distance $\rho_i$, and the ordering information is injected by the sinusoidal positional encoding~\cite{attention}. 
Thus, the node embedding is computed as 
\begin{equation*}
    \bm{h}_i = W\bm{x}_i + \bm{b} + \text{PE}(\bm{x}_i),
    \label{PE}
\end{equation*}
where $\text{PE}$ denotes the sinusoidal positional encoding. 

Given node embeddings $\{\bm{h}_i \mid n_i\in \mathcal{N}_K(c_t)\}$, a Multi-Head Attention (MHA) layer is executed to compute the node selection scores. 
The context embedding used for decoding typically relies on the embedding of the current node~\cite{POMO}, while the current node in our local encoding is the original point of the polar coordination, and its features are all zeros. 
To address this issue, we introduce a learnable vector $\bm{h}_c\in \mathbb{R}^{d_e}$ to represent the context. 
We then aggregate the information of node embeddings to the context vector using an MHA layer as follows:
\begin{equation*}
    \bm{k}_i = W^k\bm{h}_i, \bm{v}_i = W^v\bm{h}_i, \bm{q} = W^q \bm{h}_c,
    \label{qkv}
\end{equation*}
\begin{equation*}
    \tilde{\bm{h}}_c = \text{MHA}([\bm{k}_1, \dots, \bm{k}_K],[\bm{v}_1,\dots,\bm{v}_K], \bm{q}),
    \label{MHA}
\end{equation*}
where the keys $\bm{k}_i$ and values $\bm{v}_i$ are from the node embeddings, and the query $\bm{q}$ is from the context vector. The detailed computation of MHA is provided in Appendix~\ref{section:local}. Finally, the compatibility of the aggregated context vector $\tilde{\bm{h}}_c$ and the node embeddings is computed as
\begin{equation*}
    u_{\text{local}}^i = \frac{\tilde{\bm{h}}_c^{\mathrm{T}} \bm{h}_i}{\sqrt{d_e}},\;\forall n_i\in \mathcal{N}_K(c_t),
    \label{score}
\end{equation*}
which determines the action scores for selecting nodes.

\subsection{Ensemble of Global and Local Policies}
We integrate the global policy (which is implemented as POMO~\cite{POMO} in the paper) and the auxiliary local policy to generate an ensemble policy, where the strong in-distribution learning capacity of the global policy and the out-of-distribution transferability of the local policy are combined to boost generalization. 
Before integration, we adjust the original global policy by adding a distance penalty for better generalization as follows. 
\paragraph{Normalized distance penalty for global policy.} Considering that most optimal actions are included in the local neighbor nodes, 
we utilize normalized distances to penalize the global policy.
The distance penalty encourages the policy to prefer nearby nodes and be conservative to select remote nodes, which is practically helpful for generalization. 
Unlike previous methods that directly use the distance values as bias~\cite{meta-sage}, we propose to normalize distances $\rho_i$ into $[0, 1]$ by the largest $\rho_i$ in $\mathcal{N}_K(c_t)$ and add a fixed penalty $\xi$~($\xi\ge 1$) for non-neighbor nodes, i.e., 
\begin{equation*}
    \tilde{u}^i_{\text{global}} \!= \!\left\{
    \begin{aligned}
        u^i_{\text{global}} - \frac{\rho_i}{\max\{\rho_i\mid n_i\!\in\! \mathcal{N}_K(c_t)\}},& \;\text{if}\; n_i\!\in \!\mathcal{N}_K(c_t), \\
        u^i_{\text{global}} - \xi, & \;\text{otherwise},
    \end{aligned}
    \right.
    \label{penalty}
\end{equation*}
where $\mathcal{N}_K(c_t)$ contains the $K$ nearest valid neighbor nodes of the current node $c_t$, and provides a good perspective for normalization, making the distance penalty more scalable. 

To integrate the global and local policies, we first add the action scores computed from the two base policies, i.e., $\bm{u}_{\mathrm{ens}}=\tilde{\bm{u}}_{\mathrm{global}} + \bm{u}_{\mathrm{local}}$. 
After that, the action probability of the ensemble policy $\bm{\pi}_{\mathrm{ens}}$ is computed by 
\begin{align*}
    u^i_{\mathrm{masked}} &= \left\{
        \begin{aligned}
     C\cdot \tanh(\tilde{u}_{\mathrm{global}}^i\! +\! u_{\mathrm{local}}^i), & \;\text{if node $n_i$ is valid}, \\
     -\infty, & \;\text{otherwise}, \\
    \end{aligned}
    \right.\nonumber\\
    \bm{\pi}_{\mathrm{ens}} &= \text{softmax}(\bm{u}_{\mathrm{masked}}).
    \label{ensemble_1}
\end{align*}
Instead of training each base policy independently, we use a joint training method to directly optimize $\bm{\pi}_{\mathrm{ens}}$, which encourages the global policy and the local policy to work cooperatively, so that their strengths can be well combined. 
\paragraph{Training.} In practice, we first pretrain the global policy with distance penalty for $T_1$ epochs before joint training, because the state and action space of the global policy is more complex than the local one. 
At the joint training stage, we use the policy gradient method to directly train the ensemble policy $\bm{\pi}_{\mathrm{ens}}$ for $T_2$ epochs, which includes the trainable parameters $\bm{\tilde{\theta}}$ from the global policy and $\bm{\theta}$ from the local policy. 
Following POMO, we use multiple rollouts from different start nodes to get multiple trajectories in a single feedforward, and utilize the REINFORCE~\cite{reinforce} algorithm with shared baseline~\cite{POMO} to estimate the gradient of the expected return $J$. Specifically, the average reward of multiple rollouts is used as the REINFORCE baseline, and the gradient $\nabla_{\bm{\tilde{\theta}, \theta}}J(\bm{\tilde{\theta}, \theta})$ is estimated by 
\begin{equation*}
    \begin{aligned}
        & \frac{1}{N\cdot B} \sum_{i=1}^{B}\sum_{j=1}^{N} \left(R_{i,j} - \frac{1}{N}\sum_{j=1}^{N}R_{i,j}\right) \nabla_{\bm{\tilde{\theta}, \theta}} \log \bm{\pi}_{\text{ens}}(\bm{\tau}_{i,j}), 
    \end{aligned}
     \label{avg_baseline_eq}
\end{equation*}
where $N$ is the number of trajectories that is equal to the number of nodes, $B$ is the batch size, and $R_{i,j}$ is the reward of the $j$-th trajectory $\bm{\tau}_{i,j}$ on the $i$-th instance. 

\section{Experiments}
We conduct experiments on two typical VRPs, i.e., Traveling Salesman Problem~(TSP) and Capacitated VRP~(CVRP). This section presents the experimental settings and results to evaluate the performance of our method and verify the efficacy of important components. 
Our code and data are available in \href{https://github.com/lamda-bbo/ELG}{https://github.com/lamda-bbo/ELG}. 

\subsection{Experimental Settings}
We first introduce the experimental settings, including baselines, datasets, and performance metrics. 

\paragraph{Baselines.} For non-learning heuristic methods, two highly optimized solvers LKH3~\cite{LKH3} and HGS~\cite{HGS} are compared and their results on CVRPLIB are taken from the paper of HGS~\cite{HGS}. 
For learning-based methods, we collect them into several groups according to their main ideas, which are introduced as follows.
\begin{itemize}[leftmargin=0.4cm]
    \item \textbf{POMO-based methods:} POMO~\cite{POMO} is a widely-used backbone model for solving VRPs. We include POMO and its variant Sym-POMO~\cite{sym_nco} in our experiments. As the training of both POMO and Sym-POMO suffers from severe over-fitting, which results in degraded generalization performance, we early stop their training based on a validation dataset with larger scale ($N\!=\!100$ for training and $N\!=\!200$ for validation). 
    \item \textbf{Cross-distributon methods:} For those methods that improve cross-distribution generalization, we compare the state-of-the-art methods AMDKD~\cite{AMDKD} and Omni-POMO~\cite{omni-vrp}, where Omni-POMO also considers cross-scale performance. 
    \item \textbf{Cross-scale methods:} We compare three end-to-end learning methods that consider cross-scale generalization, Pointerformer~\cite{pointerformer}, LEHD~\cite{LEHD} and BQ~\cite{BQ-NCO}. For BQ, we select the best BQ-transformer model for comparison. We also compare a novel divide-and-conquer method TAM~\cite{real-time} on CVRPLIB Set-X~($N\!\ge\!500$) and Set-XXL. Since the code of TAM has not been released, and their reported results are not aligned with ours, we provide the comparison with TAM in Appendix~\ref{section:TAM} due to space limitation.
    \item \textbf{Diffusion model-based methods:} Recent works utilize the diffusion model to predict the solution distributions as heatmaps, showing great potential in generalization, especially on TSBLIB. We select DIFUSCO~\cite{difusco} and T2TCO~\cite{T2TCO} for comparison. 
\end{itemize}
All the learning methods use greedy mode at the inference stage. POMO-based methods include $\times 8$ instance augmentation technique~\cite{POMO}. Our proposed method ELG is implemented by setting the global policy as POMO, denoted as ELG-POMO. Its detailed hyperparameter settings are provided in Appendix~\ref{section:hp} due to space limitation. Note that the training of ELG-POMO is aligned with POMO; thus, the comparison between them could be seen as an ablation study of the proposed local policy.

\paragraph{Datasets.} The training instances are randomly generated with uniform node distribution and the problem scale $N=100$, which is a common setting in previous works~\cite{AM,POMO}. More details of training settings can be found in Appendix~\ref{section:training}. We consider the following test datasets to compare different methods. 
\begin{itemize}[leftmargin=0.4cm]
    \item \textbf{TSPLIB and CVRPLIB Set X \& XXL:} These benchmarks cover diverse problem instances with complex and unknown distributions together with varying and large scales, which can be used to evaluate the cross-distribution and cross-scale generalization in more realistic scenarios. 
    \item \textbf{Cross-distribution small-scale instances:} Some cross-distribution methods~(e.g., AMDKD~\cite{AMDKD}) only focus on small-scale instances. Testing them on large-scale instances like CVRPLIB may overshadow their advantages. Thus, we conduct experiments under their settings, where small-scale instances with complex synthetic distributions including cluster, cluster mixture, explosion, and rotation~\cite{vrp_distribution} are used for evaluation.
    \item \textbf{Uniform small-scale instances:} The improvement on out-of-distribution~(scale) generalization should not sacrifice the in-distribution~(scale) performance. 
    Therefore, we include the uniform small-scale dataset to examine the in-distribution~(scale) performance. 
\end{itemize}

\paragraph{Performance metrics.} We report the average gap to Best Known Solutions~(BKS), i.e., $\frac{1}{n}\sum_{i=1}^n\frac{f^*_i-f_i(\text{BKS})}{f_i(\text{BKS})}$, on each dataset to evaluate the generalization performance, where $n$ is the number of instances, $f^*_i$ and $f_i(\text{BKS})$ denote the objective value of the solution generated by each method and the BKS, respectively. We also report the average time of solving one instance on TSPLIB and CVRPLIB to measure the efficiency. 

\subsection{Results on TSPLIB and CVRPLIB}
For fair comparisons, all the learning methods are trained on uniform instances with the problem scale $N=100$, except that Omni-POMO has a stronger training setting that uses instances with complex distributions and varying scales~\cite{omni-vrp}. 
In the following, we analyze the generalization performance of different methods in detail. The CVRPLIB Set-X and TSPLIB are divided into two subsets according to the problem scales, which are $N\le200$ and $200<N\le1000$ (or $200<N\le1002$). First, we examine the cross-distribution performance on the $N\in(0,200]$ subset, and use the $N\in(200,1000]$ (or $N\in(200,1002]$) subset to compare the cross-distribution and cross-scale performance simultaneously. Then, the CVRPLIB Set-XXL is used to evaluate the performance on real-world instances with $N\in[3000,7000]$. Finally, we compare our method with highly optimized solvers LKH3 and HGS. 
\paragraph{Cross-distribution performance.} Because the test distribution is complex and the problem scales of $N\leq200$ are close to that of training instances, we can evaluate the cross-distribution performance by benchmarks with $N\leq 200$. As shown in the second columns of Tables~\ref{main_results_X} and \ref{main_results_TSPLIB}, ELG-POMO outperforms all the methods on the $N\le200$ subset, and the gap reductions on CVRP and TSP are $0.53\%$ and $0.50\%$, respectively, compared to the runner-up (Omni-POMO for CVRP, and BQ for TSP). These results clearly demonstrate the superiority of ELG-POMO in cross-distribution generalization, which will be further verified by the results in Table~\ref{cross-distribution}. 
\paragraph{Cross-distribution and cross-scale performance.} The subset of $N\in(200, 1000]$ (or $N\in(200, 1002]$) contains instances with both cross-distribution and cross-scale characteristics. ELG-POMO reduces the gap from $6.95\%$ of Omni-POMO to $6.46\%$ for CVRP, and reduces the gap from $7.47\%$ of Omni-POMO to $5.90\%$ for TSP. Considering that the Omni-POMO is trained on cross-distribution~(uniform and Gaussian mixture) and cross-scale~($N\in[50,200]$) instances while ELG-POMO only uses uniform instances with the problem scale $N=100$ in the training, the above comparison clearly demonstrates the better cross-distribution and cross-scale performance of the proposed method. Note that ELG-POMO outperforms BQ and LEHD on CVRP, but lags behind them on TSP with $N \in (200,1002]$. This may be attributed to their heavy decoder or decoder-only architectures, which can learn strong scale-invariant features~\cite{LEHD} to achieve good large-scale performance. We can also observe from Table~\ref{main_results_TSPLIB} that  ELG-POMO can surpass the diffusion mode-based methods, i.e., DIFUSCO and T2TCO, under the greedy inference setting.

\begin{table}[ht]
    \small
    \centering
    \begin{tabular}{l|c|c|c|c}
        \toprule
        Method & (0, 200] & (200, 1000] & Total & Time\\
        \midrule
        LKH3 & 0.36\% & 1.18\% & 1.00\% & 16m \\
        HGS & 0.01\% & 0.13\% & 0.11\% & 16m \\
        HGS~(less time) & 0.21\% & 1.28\% & 1.59\% & 10s \\
        \midrule
        POMO & 5.26\% & 11.82\% & 10.37\% & 0.80s\\
        Sym-POMO & 9.99\% & 27.09\% & 23.32\% & 0.87s\\
        Omni-POMO & 5.04\% & 6.95\% & 6.52\% & 0.75s\\
        LEHD & 11.11\% & 12.73\% & 12.25\% & 1.67s \\
        BQ & 10.60\% & 10.97\% & 10.89\% & 3.36s\\
        \midrule
        ELG-POMO~(Ours) & \textbf{4.51\%} & \textbf{6.46\%} & \textbf{6.03\%} & 1.90s\\        
        \bottomrule     
    \end{tabular}
    \caption{Empirical results on CVRPLIB Set-X covering cross-distribution and cross-scale test instances. Note that Pointerformer, DIFUSCO, and T2TCO are not intended for CVRP. }
    \label{main_results_X}\vspace{-0.5em}
\end{table}

\begin{table}[ht]
    \small
    \centering
    \begin{tabular}{l|c|c|c|c}
        \toprule
        Method & (0, 200] & (200, 1002] & Total & Time\\
        \midrule
        LKH3 & 0.00\% & 0.00\% & 0.00\% & 24s \\
        \midrule
        POMO & 3.07\% & 13.35\% & 7.45\% & 0.41s\\
        Sym-POMO & 2.88\% & 15.35\% & 8.29\% & 0.34s \\
        Omni-POMO & 1.74\% & 7.47\% & 4.16\% & 0.34s\\
        Pointerformer & 2.31\% & 11.47\% & 6.32\% & 0.24s \\
        LEHD & 2.03\% & 3.12\% & 2.50\% & 1.28s\\
        BQ & 1.62\% & \textbf{2.39\%} & \textbf{2.22\%} & 2.85s\\
        DIFUSCO & 1.84\% & 10.83\% & 5.77\% & 30.68s\\
        T2TCO & 1.87\% & 9.72\% & 5.30\% & 30.82s\\
        \midrule
        ELG-POMO~(Ours) & \textbf{1.12\%} & 5.90\% & 3.08\% & 0.63s\\
        \bottomrule     
    \end{tabular}
    \caption{Empirical results on TSPLIB covering cross-distribution and cross-scale test instances.}
    \label{main_results_TSPLIB}\vspace{-0.5em}
\end{table}

\paragraph{Performance on real-world CVRP instances.} We evaluate our method on four real-world instances, i.e., Antwerp1~(A1), Antwerp2~(A2), Leuven1~(L1), and Leuven2~(L2), with $N\in[3000, 7000]$ from CVRPLIB Set-XXL.
As shown in Table~\ref{cvrplib-xxl}, the ELG-POMO method performs the best on three instances and significantly reduces the optimality gap compared to other POMO-based methods. 

\begin{table}[ht]
    \small
    \centering
    \resizebox{.99\columnwidth}{!}{
    \begin{tabular}{l|c|c|c|c}
        \toprule
        Method & A1 & A2 & L1 & L2 \\
        \midrule
        POMO & 112.27\% & 159.22\% & 75.30\% & 78.16\%  \\
        Sym-POMO & 79.72\% &  179.55\% & 170.44\% & 179.55\% \\
        Omni-POMO & 42.52\% & 48.59\% & 22.79\% & 60.39\% \\
        LEHD & 18.90\% & 26.40\% & 14.04\% & 26.30\% \\
        BQ & 11.21\% & \textbf{15.02\%} & 13.27\% & 24.00\% \\
        \midrule
        ELG-POMO~(Ours) & \textbf{10.70\%} & 17.69\% & \textbf{10.77\%} & \textbf{21.80\%} \\
        \bottomrule
    \end{tabular}}
    \caption{Empirical results on four real-world instances from CVRPLIB Set-XXL with $N\in[3000,7000]$. }
   \label{cvrplib-xxl}
\end{table}

\paragraph{Comparison to non-learning heuristic solvers.} We can observe from Table~\ref{main_results_X} that the time consumption of ELG-POMO is smaller than LKH3 and HGS (1.90s\:vs.\:16m). However, its optimality gap is worse, even when HGS uses less time, implying that further improvement needs to be done for neural construction methods in the future. 
\paragraph{Runtime.} As shown in Tables~\ref{main_results_X} and \ref{main_results_TSPLIB}, ELG-POMO costs a lower runtime than BQ and diffusion model-based methods. Compared to the pure POMO method, introducing an additional local policy increases the time cost, but is acceptable. 

\subsection{Results on Synthetic Distributions}
We also conduct experiments on small-scale instances generated from synthetic distributions. First, we will compare ELG-POMO with AMDKD on small-scale instances from four complex distributions to further verify the superiority of ELG-POMO in cross-distribution generalization. Second, we will show that the proposed ELG-POMO does not worsen the in-distribution~(scale) performance when improving the out-of-distribution~(scale) performance. In this subsection, we report the gap of average objective values following the common setting~\cite{AM,AMDKD}. 

\paragraph{Cross-distribution performance.} AMDKD is the state-of-the-art method for cross-distribution generalization~\cite{AMDKD}. As it is not designed for cross-scale generalization, we present the comparison with AMDKD here. For fairness, we retrain the POMO and ELG-POMO~(denoted as POMO* and ELG*) using three exampler distributions: uniform, cluster, and cluster mixture, which are aligned with AMDKD. Then we select four difficult distributions to evaluate them. The results on TSP are shown in Table~\ref{cross-distribution}, where ELG* outperforms AMDKD on three distributions with $N=100$ and four distributions with $N=200$. The gap reductions on four distributions with $N=200$ are roughly $1.0\%$--$1.9\%$, which are very significant. Similar experimental results on CVRP are presented in Appendix~\ref{section:CVRP_cross_dis} due to space limitation. 
\begin{table}[ht]
    \small
    \centering
    \begin{tabular}{l|c|c|c|c}
        \toprule
        Distributions & Scale & POMO* & AMDKD & ELG*\\
        \midrule
        Cluster & $N=100$ & 0.45\% & \textbf{0.35\%} & 0.36\% \\
        Mixture & $N=100$ & 0.49\% & 0.41\% & \textbf{0.37\%} \\
        Explosion & $N=100$ & 0.29\% & 0.30\% & \textbf{0.23\%} \\
        Rotation &$N=100$ & 0.38\% & 0.32\% & \textbf{0.31\%} \\
        \midrule
        Cluster & $N=200$ & 2.63\% & 3.01\% & \textbf{1.84\%} \\
        Mixture & $N=200$ & 2.45\% & 3.50\% & \textbf{1.62\%}
        \\
        Explosion &$N=200$ & 2.18\% & 2.63\% & \textbf{1.43\%} \\
        Rotation & $N=200$ & 2.33\% & 2.54\% & \textbf{1.58\%} \\
        \bottomrule
    \end{tabular}
    \caption{Empirical results on TSP instances with four difficult synthetic distributions. }
    \label{cross-distribution}\vspace{-0.5em}
\end{table}
\paragraph{In-distribution test performance.} We train and test all the methods on uniform instances with $N=100$. The results are shown in Table~\ref{iid}, where we use the solutions generated by Concorde~\cite{concorde} and LKH3 as the BKS for TSP and CVRP, respectively, to compute gaps. For TSP100, the proposed ELG-POMO achieves competitive performance to state-of-the-art methods for TSP~(Pointerformer and T2TCO) with $0.04\%$--$0.06\%$ difference in the gap. For CVRP100, ELG-POMO surpasses all the compared methods. Note that although BQ and LEHD improve the cross-distribution and cross-scale generalization as shown in Tables~\ref{main_results_X} to~\ref{cvrplib-xxl}, their in-distribution test results get worse than POMO, except for BQ's result on TSP100. On the contrary, besides the clear advantage of cross-distribution and cross-scale generalization, the proposed ELG-POMO can improve the in-distribution performance to some extent, i.e., the introduction of transferrable local policy also benefits the generalization to unseen in-distribution~(scale) instances. 
\begin{table}[ht]
    \small
    \centering
    \begin{tabular}{l|c|c}
        \toprule
        Method & TSP100 & CVRP100\\
        \midrule
        Concorde & 0.00\% & -- \\
        LKH3 & 0.00\%  & 0.00\%  \\
        HGS & -- & -0.52\%  \\
        \midrule
        POMO & 0.40\% & 1.34\% \\
        Sym-POMO & 0.41\%& 1.43\% \\
        Pointerformer & \textbf{0.16\%} & -- \\
        LEHD & 0.57\% & 3.64\% \\
        BQ & 0.35\% & 2.76\%\\
        DIFUSCO & 0.24\% & --  \\
        T2TCO & 0.18\% & -- \\
        \midrule
        ELG-POMO~(Ours) & 0.22\% &\textbf{1.22\%} \\
        \bottomrule
    \end{tabular}
    \caption{Empirical results on uniform instances with $N=100$, where ``--" means that the method is not intended for that problem.}
    \label{iid}\vspace{-0.5em}
\end{table}
\subsection{Ablation Studies}
We conduct ablation studies on three components associated with the local policy: normalized distance penalty, positional encoding, and polar coordinates. The results show that these components really contribute to improving performance, and the proposed ELG method still achieves good generalization performance even without these components. Detailed results are provided in Appendix~\ref{section:ablation1} due to space limitation. 

We also examine the influence of two new hyperparameters introduced by the proposed method: the fixed penalty $\xi$ and the local neighborhood size $K$. The results show that when $\xi$ and $K$ are within reasonable ranges, they do not affect the performance significantly. 
Detailed results are provided in Appendix~\ref{section:ablation2} due to space limitation. 
\section{Conclusion}
The generalization issue of NCO methods has prevented them from real-world applications. In this paper, we propose to ensemble a global policy and a transferrable local policy for VRPs, where the global policy learns from the global information of a complete VRP graph, resulting in its strong in-distribution learning capacity, and the local policy learns from the local topological features which are insensitive to the variation of node distribution and problem scale, leading to its good transferability. The combination of these two complementary policies leads to better generalization, which is verified by the extensive comparison with state-of-the-art methods on two well-known benchmarks (i.e., TSPLIB and CVRPLIB) covering diverse problem instances. A limitation of the proposed method is that it costs more time to solve very large-scale instances than divide-and-conquer methods as shown in Appendix~\ref{section:TAM}. In future work, we will try to further improve the latency and extend the proposed method to more CO problems, such as the pickup-and-delivery problem. 

\section*{Acknowledgments}
The authors want to thank the anonymous reviewers for
their helpful comments and suggestions. This work was
supported by the National Science and Technology Major Project (2022ZD0116600) and National Science Foundation of
China (62276124). Chao Qian is the corresponding author. The conference version of this paper has appeared at IJCAI’24.

\bibliographystyle{named}
\bibliography{ijcai24}

\newpage
\appendix
\section{Detailed Hyperparameter Settings}
\label{section:hp} 
The detailed hyperparameter settings of our method are listed in Table~\ref{hp}. The configuration of global policy follows the default setting of POMO~\cite{POMO}. The number of training epochs $T_1 + T_2$ is aligned with the early stopping point of POMO to alleviate the overfitting of the global policy, which are $35$ epochs for TSP and $25$ epochs for CVRP. Concretely, these early stopping points are determined by manually observing the performance curve on the validation dataset with $N=200$. Moreover, the sensitivity of newly introduced parameters, i.e., fixed penalty $\xi$ and local neighborhood size $K$, is verified in Appendix~\ref{section:ablation2}. The scaling parameter $C$ is increased to 50, which can speed up the convergence~\cite{pointerformer}. 
\begin{table}[ht]
    \small
    \centering
    \resizebox{.95\columnwidth}{!}{
    \begin{tabular}{l|c|c}
        \toprule
        Hyperparameter & For solving TSP & For solving CVRP\\
        \midrule
        $T_1$ & 30 & 20\\
        $T_2$ & 5 & 5 \\
        epoch size & 1.2M & 1.2M \\
        batch size & 120 & 120 \\
        learning rate & 1e-4 & 1e-4 \\
        weight decay & 1e-6 & 1e-6 \\
        \midrule
        scaling parameter $C$ & 50 & 50 \\
        fixed penalty $\xi$ & 1.0 & 1.0 \\
        local neighborhood size $K$ & 30 & 40\\
        \midrule
        \multicolumn{3}{l}{Configuration of attention-based local policy} \\
        \midrule
        embedding dimension $d_e$ & 32 & 32 \\
        number of heads & 4 & 4 \\
        dimension of single head & 8 & 8 \\
        \bottomrule
    \end{tabular}}
    \caption{Hyperparameters of our method for solving TSP and CVRP. }
    \label{hp}\vspace{-0.5em}
\end{table}
\section{Additional Implementation Details}
\subsection{Implementation Details of Local Policy}
\label{section:local}
\paragraph{Multi-head attention for decoding.}
Section~\ref{section:local_net} introduces the general process of using Multi-Head Attention~(MHA) to decode action scores. We present the detailed computation process of MHA here. First, we reshape the query $\bm{q}\in \mathbb{R}^{d_k\times M}$ of context, the keys and values $\bm{k}_i,\bm{v}_i\in \mathbb{R}^{d_k\times M}$ of node $n_i$ into $M$ parts for multi-head computation, where $M$ is the number of heads and $d_k$ is the embedding dimension of single-head attention. For each head $j\in [M]$, where $[M]$ denotes the set $\{1,2,\ldots,M\}$, we compute the compatibility $u_{i,j}\in \mathbb{R}$ of the single-head query $\bm{q}_j\in \mathbb{R}^{d_k}$ with the single-head key $\bm{k}_{i,j}\in \mathbb{R}^{d_k}$ of node $n_i$ as the scaled dot-product: 
\begin{equation*}
    u_{i,j} = \frac{{\bm{q}}^{\mathrm{T}}_j \bm{k}_{i,j}}{\sqrt{d_k}},\;\forall n_i\in \mathcal{N}_K(c_t).
\end{equation*}
Based on the compatibility, we compute the attention weights $\alpha_{i,j}\in [0,1]$ using a softmax: 
\begin{equation*}
    \alpha_{i,j} = \frac{e^{u_{i,j}}}{\sum_{i}e^{u_{i,j}}}.
\end{equation*}
The messages $\bm{v}_{i,j}$ of node $n_i$ are then aggregated to context vector:
\begin{equation*}
    \bm{h}_c^j = \sum_i \alpha_{i,j}\bm{v}_{i,j}.
\end{equation*}
Finally, we use a linear layer to project $\bm{h}_c^j$ for $j\in[M]$ back to a single $d_e$-dimension vector using parameter matrices $W^O_j\in \mathbb{R}^{d_e\times d_k}$: 
\begin{equation*}
    \tilde{\bm{h}}_c = \sum_{j=1}^M W^O_j \bm{h}_c^j.
\end{equation*}
\paragraph{Padding.} Considering that the number of valid nodes may not be the same when solving a batch of instances, we pad the local state $\bm{s}_t$ to the max length by adding zeros. 
The max length is computed by $\min\{K, n_{\text{max}}\}$, where $K$ is the local neighborhood size and $n_{\text{max}}$ is the max number of valid nodes in the batch of instances. Then, an attention mask is used to mask out the padded zeros. 
\subsection{Implementation Details in Training} 
\label{section:training}
Following the common setting to generate training instances~\cite{AM,POMO}, the node coordinates $\{(x_i,y_i)|\;i\in [N]\}$ are randomly generated from a uniform distribution $\mathcal{U}(0,1)$, the capacity $Q$ is set to $50$, and the demands $\{d_i|\;i\in[N]\}$ are randomly generated from a uniform distribution $\mathcal{U}(1,10)$. 

For comparison to AMDKD, we retrain POMO and ELG-POMO on three node distributions: Uniform~(U), Cluster~(C), and cluster Mixture~(M), which are generated according to the code of AMDKD~\cite{AMDKD}. We also use an adaptive multi-distribution scheduler similar to AMDKD. That is, we define that the probability of selecting a distribution $d\in\{\text{U}, \text{C}, \text{M}\}$ is proportional to the exponent value of the gap on the validation dataset with the corresponding $d$. 

Because the range of objective function values may be different across instances, especially training on cross-distribution or cross-scale datasets, we conduct a normalization on the advantage terms in REINFORCE as  
\begin{equation*}
    \begin{aligned}
        A_i & = R_i - \frac{1}{N}\sum_{i=1}^{N}R_i,\ i\in [N],\\
        \tilde{A}_i &  = A_i / \max \{A_i|\;i\in [N]\}. \\
    \end{aligned}
    \label{norm_advantage}
\end{equation*}

\section{Additional Results}\label{section:additional results}
\subsection{Ablation Study of Components}
\label{section:ablation1}
We conduct an ablation study on three components associated with the local policy, which are normalized distance penalty~(denoted as Penalty), positional encoding~(denoted as PE), and polar coordinates~(denoted as Polar). 
The experimental results are shown in Table~\ref{ablation}. We illustrate the role of each component as follows. 
\paragraph{Normalized distance penalty.} We propose to normalize the distance penalty by the largest distance within the local neighborhood $\mathcal{N}_K(c_t)$ to make the penalty more scalable. By comparing the results in the 2nd and 3rd columns of Table~\ref{ablation}, we can find that the normalized distance penalty contributes a lot to both in-distribution (scale) performance on TSP100 and CVRP100, as well as out-of-distribution (scale) performance on TSPLIB and CVRPLIB (X). 
The penalty reduces the gap by roughly $1.10\%$ on TSPLIB and $1.15\%$ on CVRPLIB (X). Note that ELG w/o penalty still outperforms most baseline methods in Tables~\ref{main_results_X} and \ref{main_results_TSPLIB}. 
\begin{table}[ht]
    \small
    \centering
    \resizebox{.98\columnwidth}{!}{
    \begin{tabular}{l|c|c|c|c}
        \toprule
        Method & ELG & w/o Penalty & w/o PE & w/o Polar\\
        \midrule
        TSP100 & 0.22\% & 0.39\% (-) & 0.24\% (-) & 0.23\% (-) \\
        CVRP100 & 1.22\% & 1.32\% (-) & 1.25\% (-) & 1.24\% (-) \\
        \midrule
        TSPLIB & 3.08\% & 4.19\% (-) & 3.34\% (-) & 3.58\% (-)\\
        CVRPLIB (X) & 6.03\%  & 7.19\% (-) & 5.97\% (+) & 6.52\% (-)\\
        \bottomrule
    \end{tabular}}
    \caption{Emprical results of ablation study, where w/o Penalty denotes the ELG method without the normalized distance penalty, w/o PE denotes the ELG method without positional encoding, and w/o Polar denotes the ELG method using Euclidean coordinates instead of polar ones. ``-" and ``+" denote that the performance deteriorates and improves, perspectively. }
    \label{ablation}\vspace{-0.5em}
\end{table}
\paragraph{Positional encoding.} We use positional encoding to incorporate distance ordering information into the embeddings, which can benefit the local node selection. By comparing the results in the 2nd and 4th columns, PE can improve the performance on three test datasets. 
\paragraph{Polar coordinates.} Since we use a lightweight network with only two layers to parameterize the local policy, polar coordinates that explicitly provide the distance information are easier to learn. By comparing the results in the 2nd and 5th columns, we find that using Euclidean coordinates to replace the polar ones will degrade the performance in all cases. 

\subsection{Sensitivity Analysis of Hyperparemeters}
\label{section:ablation2}
The ELG method introduces two new hyperparameters: the fixed penalty $\xi$ and the local neighborhood size $K$. 

As for the fixed penalty $\xi$, it should be greater than $1$ so that the penalty for non-neighbor nodes is larger than that for neighbor nodes, but $\xi$ should not be too large to mask non-neighbor nodes, since they can still be involved in the optimal solutions. 
We compare the performance of using $\xi = 1.0, 1.5,$ and $2.0$ in Table~\ref{xi}. The results show that $\xi$ does not affect the performance significantly. 

\begin{table}[ht]
    \small
    \centering
    \begin{tabular}{l|c|c}
        \toprule
        $\xi$ & TSP100 & TSPLIB\\
        \midrule
        1.0 & 0.22\% &  3.08\%\\
        1.5 & 0.22\% & 3.41\% \\
        2.0 & 0.22\% & 3.31\% \\
        \bottomrule
    \end{tabular}
    \caption{Emprical results of ELG-POMO using different $\xi$.}
    \label{xi}\vspace{-0.5em}
\end{table}

Table 9 shows the results of ELG-POMO using the local neighborhood size $K\in \{20, 30, 40, 50\}$. We can observe that $K$ does not influence the performance significantly, and even the results of the worst $K$ are clearly better than that of POMO~($0.40\%$ on TSP100 and $7.45\%$ on TSPLIB). 

\begin{table}[ht]
    \small
    \centering
    \begin{tabular}{l|c|c}
        \toprule
        $K$ & TSP100 & TSPLIB\\
        \midrule
        20 & 0.22\% & 3.22\%\\
        30 & 0.22\% & 3.08\% \\
        40  & 0.22\% & 3.51\%\\
        50 & 0.23\% & 3.57\% \\
        \bottomrule
    \end{tabular}
    \caption{Emprical results of ELG-POMO using different $K$. }
    \label{K}\vspace{-0.5em}
\end{table}

\subsection{Comparison with TAM}
\label{section:TAM}
TAM is a divide-and-conquer method that aims to solve large-scale instances and can be combined with both heuristic solvers and learning-based methods. Because TAM's paper~\cite{real-time} only reported the results on instances of CVRPLIB Set-X and Set-XXL with $N\ge500$, which are not aligned with our main results in Tables~\ref{main_results_X} and~\ref{cvrplib-xxl}, we put the comparison here to further verify the performance on large-scale instances. We compare ELG-POMO with the combination of TAM and LKH3, denoted as TAM-LKH3. 
As shown in Table~\ref{tam}, ELG-POMO consistently achieves better optimality gaps on both Set-X~($2.07\%$ improvement) and Set-XXL~($5.20\%$ improvement). However, the runtime of ELG-POMO is longer than TAM-LKH3 on Set-XXL instances with very large scales, which may be due to the quadratic time complexity of the Transformer architecture.  

\begin{table}[ht]
    \small
    \centering
    \begin{tabular}{l|c|c}
    \toprule
    Metrics/ Methods & TAM-LKH3 & ELG-POMO \\
    \midrule
    Set-X~($N\ge500$) & 9.87\% & 7.80\% \\
    Avg time on Set-X & 2.45s & 1.90s\\
    Set-XXL~($N\in[3000,7000]$) & 20.44\% & 15.24\%\\
    Avg time on Set-XXL & 24.03s & 190.45s\\
    \bottomrule
    \end{tabular}
    \caption{Emprical results of comparisons to the divide-and-conquer method TAM-LKH3. }
    \label{tam}
\end{table}

\subsection{Comparison on CVRP Instances with Complex Distributions}
\label{section:CVRP_cross_dis}
We compare the proposed method with AMDKD on CVRP instances with four complex distributions. Due to space limitation of the main paper, we present the results here. We retrain POMO and ELG-POMO~(denoted as POMO* and ELG* in Table~\ref{cross-distribution-cvrp}) with three exampler distributions, i.e., uniform, cluster, and cluster mixture. When training on these datasets, we increase $T_1$ from $20$ to $40$ so that POMO and ELG-POMO can achieve better convergence on small-scale CVRP instances than early stopping. Note that increasing $T_1$ to $40$ does not make the number of training instances of ELG-POMO larger than that of AMDKD. The results in Table~\ref{cross-distribution-cvrp} show that ELG-POMO (i.e., ELG*) always outperforms AMDKD. 
\begin{table}[ht]
    \small
    \centering
    \begin{tabular}{l|c|c|c|c}
        \toprule
        Distributions & Scale & POMO* & AMDKD & ELG*\\
        \midrule
        Cluster & $N=100$ & 1.44\% & 1.36\% & \textbf{1.34\%} \\
        Mixture & $N=100$ & 1.12\% & \textbf{0.99\%} & 1.02\% \\
        Explosion & $N=100$ & 1.48\% & 2.01\% & \textbf{1.35\%} \\
        Rotation &$N=100$ & 1.39\% & 1.99\% & \textbf{1.32\%} \\
        \midrule
        Cluster &$N=200$ & 3.15\% & 8.62\% & \textbf{2.96\%} \\
        Mixture &$N=200$ & 2.72\% & 7.36\% & \textbf{2.40\%} \\
        Explosion &$N=200$ & 3.02\% & 6.61\% & \textbf{2.81\%} \\
        Rotation & $N=200$ & 2.83\% & 6.85\% & \textbf{2.71\%} \\
        \bottomrule
    \end{tabular}
    \caption{Empirical results on CVRP instances with difficult synthetic distributions. }
    \label{cross-distribution-cvrp}\vspace{-0.5em}
\end{table}

\subsection{Results of only local policy}
The transferrable local policy is a critical component of our proposed method, which contributes greatly to both cross-distribution and cross-scale performance. Therefore, we evaluate the only local policy for clear ablation. The empirical results are shown in Table~\ref{only_local}. For comparison fairness, we train a local policy with a similar number of parameters as the global policy by adding $3$ encoder layers and increasing the embedding dimension to $128$. 

By comparing the results in the 2nd and 3rd rows, we observe that the local policy significantly improves the out-of-distribution performance on TSPLIB, verifying the good transferability of local policy. However, the local policy's in-distribution performance on TSP100 deteriorates, probably because of the reduced state and action spaces. 

As shown in Table~\ref{only_local}, neither the local policy nor global policy performs worse than the proposed ELG-POMO method and these two policies exhibit complementary performances, validating the necessity of combining the local and global policies in our approach. 

\begin{table}[ht]
    \small
    \centering
    \begin{tabular}{l|c|c}
        \toprule
        Methods & TSP100 & TSPLIB\\
        \midrule
        Only local policy & 1.71\% & 3.81\%\\
        Only global policy (POMO) & 0.40\% & 7.45\% \\
        ELG-POMO  & 0.22\% & 3.08\%\\
        \bottomrule
    \end{tabular}
    \caption{Ablation results of only the local policy. }
    \label{only_local}\vspace{-0.5em}
\end{table}

\section{Test Datasets and Computational Devices}
\paragraph{Test Dataset.} We present some detailed information of test datasets here. For TSPLIB, we select a subset with the number of nodes $N\le 1002$ to evaluate the performance, which contains $49$ instances. For CVRPLIB Set-XXL, we select a subset of $4$ instances with $N\le 10000$. For CVRPLIB Set-X, we use all the $100$ instances with $N\le 1000$. Note that TSPLIB and CVRPLIB Set-XXL both contain real-world instances with complex node distributions and large scales, and the instances of CVRPLIB Set-X \& XXL also have varying capacities and complex demand distributions, which are quite hard to generalize. These TSPLIB and CVRPLIB datasets can be obtained from \cite{TSPLIB,vrplib_x,real-world}. Other synthetic datasets are randomly generated using a fixed random seed $1234$. The generation of uniform datasets, TSP100 and CVRP100, follows the code of \cite{AM}. The generation of cluster and cluster mixture datasets follows the code of \cite{AMDKD}, and the generation of explosion and rotation datasets follows the code of \cite{vrp_distribution}. TSP100 and CVRP100 both contain $10000$ instances and other datasets with complex distributions contain $1000$ instances, which follows the common setting~\cite{AM,AMDKD}. 
\paragraph{Computational devices.} We train our models on one GPU~(NVIDIA RTX 6000 Ada) with $48$GB memory.  We test all the learning-based methods on one NVIDIA GeForce RTX 4090 and record their time consumption. For non-learning methods, we run them on the CPU~(AMD EPYC 7763 64 cores) and record their time consumption.  
\end{document}